\documentclass[11pt,a4paper]{article}
\usepackage[hyperref]{emnlp2018}
\usepackage{times}
\usepackage{latexsym}

\usepackage{url}

\usepackage{amssymb,mathtools}
\newcommand{\todo}[1]{}
\renewcommand{\todo}[1]{{\color{red} TODO: {#1}}}
\usepackage[pdftex]{graphicx}

\aclfinalcopy % Uncomment this line for the final submission

\title{Lightweight Adaptive Mixture of Neural and N-gram Language Models}

\author{Anton Bakhtin \; Arthur Szlam \; Marc'Aurelio Ranzato \; Edouard Grave \\
  Facebook AI Research, NY \\
  {\tt yolo@fb.com}  \\}

\date{}

\begin{document}
\maketitle
\begin{abstract}
It is often the case that the best performing language model is an ensemble of a neural language model with n-grams.
In this work, we propose a method to improve how these two models are combined. By using a small network
which predicts the mixture weight between the two models, we adapt  their relative importance at \textit{each} time step.
Because the gating network is small, it trains quickly on small amounts of held out data, and does not add overhead at scoring time.
Our experiments carried out on the One Billion Word benchmark show a significant improvement over the state of the art ensemble without retraining of the basic modules.
\end{abstract}

\section{Introduction}
\begin{figure}
\includegraphics[scale=0.9]{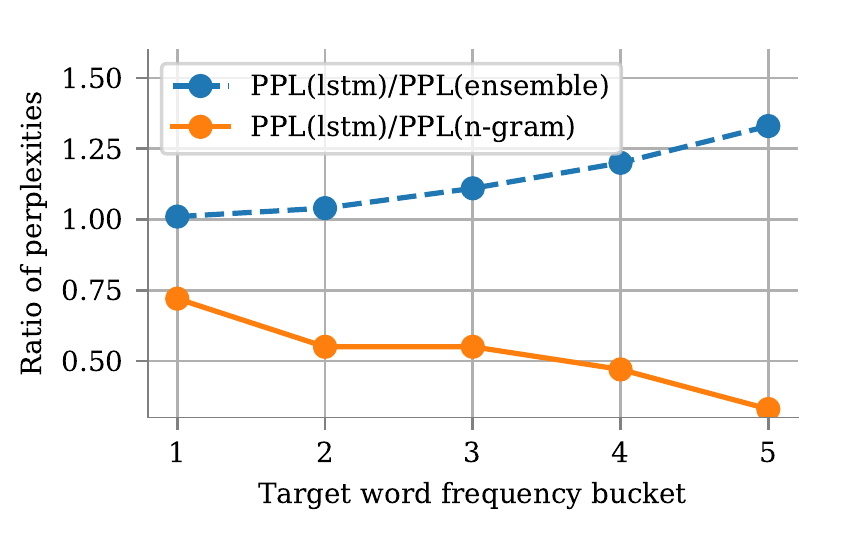}
\vspace{-1.1cm}
    \caption{ As the frequency of the word-to-predict decreases (from left to right), the relative performance of neural models
gets better compared to n-grams (orange curve). Yet, ensembling
the two models (with a fixed scalar weight) is more effective on rarer words (blue curve). Bins were built by sorting words by frequency and by dividing them into buckets with equal probability mass.}
    \label{fig:rare_words}
\end{figure}

The goal of statistical language modeling is to estimate the probability of sentences or sequences of words~\citep{asr}. By the chain rule of probability theory, this is equivalent to estimation of the conditional probability of a word given all preceding words.
This problem is key to natural language processing, with applications not only in type-ahead systems, but also machine translation~\citep{pbsmt} and automatic speech recognition~\citep{asr}.
While earlier work on statistical language modeling focused on n-gram language models~\citep{kneser1995improved, chen1999empirical}, recent advances are based on variants of neural language models~\citep{bengio2003neural,mikolov2010recurrent,dauphin2016language}, which have yielded state of the art performance on several large scale benchmarks~\cite{jozefowicz2016exploring}.
Neural approaches require less memory than n-grams and they generalize better, but with a substantial increase in computational complexity both at training and test time.
Despite the superior performance of neural models, even better results in terms of perplexity can be achieved by ensembling neural models with n-grams~\citep{mikolov2011strategies,chelba2013one,jozefowicz2016exploring}. %Ensembling is the most na\"ive way to combine the two models, by just using a single constant scalar to weigh the two predictive distributions over the next word in the sequence.
However, Fig.~\ref{fig:rare_words}, which shows results using a single constant scalar to weigh the output distribution of a neural model and an n-gram model, suggests that the relative contributions of the two models are not simple.  %non-linear, and somewhat counter-intuitive, as
For example, the neural model generalizes better than the n-gram on rarer words, yet on rarer words the ensemble yields the largest gains. %This motivates our investigation of better approaches to combine these two models.

In this work we will study more sophisticated methods for combining the results of n-gram and neural language models than a fixed scalar weight. We will propose a simple gating network which takes as input a handful of features based on frequency statistics to produce
as output an \textit{input dependent} weight to be used in the ensemble, effectively turning the ensemble into an adaptive mixture of experts model.  We show that given already trained neural and n-gram language models, the gating network can be trained quickly on a handful of examples.  The gating network consistently yields better results
than the ensemble which uses a fixed weight in the mixture, while adding a negligible computational cost.
We evaluated our proposed approach on the One Billion Word benchmark~\citep{chelba2013one}, the biggest publicly available benchmark for language modeling, and on the Wall Street Journal corpus, demonstrating seizable gains on both datasets.

\section{Related work}
\label{sec:related}
The method we propose is a particular instance of a \textit{mixture of experts} (MoEs)~\citep{jacobs1991adaptive}, where experts are pre-trained and the gating is a possibly recurrent
function of some handcrafted features. The advantages are twofold. First, we do not need to update the experts which are very large systems and instead,
we can learn very quickly to modulate between them. Second, the gating network is tiny as we do not need to represent the actual input words, which further speeds up training time and reduces sample complexity.

\citet{shazeer2017outrageously} also proposed to use a MoEs for language modeling. The major technical difference is that they employ MoEs in between layers of a deep  LSTM, while we do it at the output. While our focus is to design a lightweight system that optimally combines pre-trained models, their focus is to train a single much higher capacity system, a much more engineering involved endeavor.

In \citep{kneser1993dynamic} information about ground truth performance of several n-gram models on previous 400 words is used to predict the optimal interpolation weights at each position. However, this approach could only be applied if the ground truth is known in advance and enough context is given.

A few works explored using n-gram features within a neural language model.
\citet{mikolov2011strategies} and \citet{chelba2013one} train a neural model jointly with a maximum entropy model taking as input n-gram features.
\citep{neubig2016generalizing} proposes an approach more similar to ours, except that the gating network takes as input the hidden state of the neural model. This has two drawbacks. First, the hidden state may already have lost the discriminative information necessary for the selection of the expert. Second, the gating network operates on a much higher dimensional input, and therefore, it requires more data to train. Moreover, we do not attempt at tuning the experts nor we care about how these were trained~\citep{jean2014using, grave2016efficient}, but we use them as black-boxes and only train the gating network which is a much simpler task.

\section{Basic Models}
\label{sec:experts}
The goal of language modeling is to estimate the probability $P(w_t|\mathbf{c})$ of a next word $w_t$ given its context sequence $\mathbf{c}=(w_{t-1}, \dots, w_1)$; the context being empty if $t=1$.
In this section, we introduce the models we use to instantiate our experts.

\subsection{N-gram Language Models}
\label{sec:ngram}
N-gram models rely on the following Markov assumption: the next word depends only on the $N - 1$ previous words: $\textrm{P}(w_t | \mathbf{c}) = \textrm{P}(w_t | \mathbf{c}_{N-1})$, where $\mathbf{c}_{N-1}=(w_{t-1}, \dots, w_{t-N+1})$; maximum likelihood estimation then yields:
$\textrm{P}(w_t | \mathbf{c}_{N-1}) = C(w_t,\mathbf{c}_{N-1}) / C(\mathbf{c}_{N-1})$,
where $C(\bullet)$ stands for the number of occurrences of the sequence in the training corpus.

For high order models, e.g, $N=5$, only a small fraction of the n-grams appear in the training corpus, a problem also referred to as \textit{data sparsity},
which would yield $0$ probability for almost all sentences. To counteract this, several \textit{back-off} techniques have been suggested, the most popular being defined as:
\begin{equation}
 \label{eq:smoothed_ngram}
 \textrm{P}^{NG}_N(w_t) =
  \begin{cases}
    p_{w_t,\mathbf{c}_{N-1}} & \text{if not zero} \\
    \textrm{P}^{NG}_{N-1}(w_t) \alpha_{\mathbf{c}_{N-1}} & \text{otherwise}
  \end{cases}
\end{equation}
where $\alpha$ and $p$ are called back-off coefficients and discounted probabilities, respectively.
In this work, we use Kneser-Ney formulation~\citep{kneser1995improved} that yields state of the art results among n-gram models.

\subsection{Neural Language Model}
Another approach to reduce sparsity is to encode the context $\mathbf{c}$ as a fixed length dense vector $h_t$.
To do so each word $w$ is mapped to an embedding vector $v(w)$. The sequence of vectors $v(w_1), \dots, v(w_t)$ is then fed to a neural network $f$ to produce $h_t$.
A linear classifier $a$ is then applied to $h_t$ to estimate the probability distribution over the next word:
\begin{equation}\label{eqn:nn_lm}
\textrm{P}^{NN}(\bullet | \mathbf{c}) = a\left(f(v(w_1), \dots, v(w_{t-1}))\right).
\end{equation}
Different types of networks could be used as encoders, such as fully connected~\citep{bengio2003neural}, convolutional~\citep{dauphin2016language} or recurrent~\citep{mikolov2010recurrent,chelba2013one}.
In this work we use LSTMs~\citep{lstm}  %ADDAFTER~\citep{lstm}
which is nowadays one of the strongest performing methods.

\section{Mixture of experts}
\label{sec:main}
Different experts have different strengths and weaknesses. The complementarity of neural and n-gram language models explains why ensembling works so well.
However, it is conceivable that different contexts may need different weighting. MoEs address exactly this issue, enhancing the model with a gating network that
weighs experts in an input dependent manner. Next, we first analyze where n-grams outperform neural language models, and then propose a simple gating mechanism to automatically select the most suitable expert.

\subsection{Analysis}
\begin{table}[t]
\centering
\begin{tabular}{p{7cm}}
\dots will be made at an event at the San Francisco Museum of \textbf{Science} \\
\hline
Robert Jew of the National Archives and Records \textbf{Administration} \\
\hline
\dots We need professional advice , said Senator George H. \textbf{Winner} \\
\hline
\dots he shows up armed to buy machine guns and \textbf{siliences} \\
\hline
Xbox 360 ( R ) video game and entertainment system  \textbf{from}
\end{tabular}
\label{tab:examples_ngram_better}
\vspace{-.2cm}
\caption{ Examples of contexts where the n-gram model significantly outperforms the neural model. The word to be predicted is marked in bold font.}
\end{table}

N-gram language models require a large memory which grows with the amount of training data~\cite{silva2016theoretical}, but are fast at test time as they require only table lookups. They can easily
memorize patterns but do not generalize well to rare events.
On the other hand, neural language models are much more compact, generalize much better but require more computation.
\citet{jozefowicz2016exploring} found that the relative advantage of neural models increases as the frequency of target word decreases.
While we observe the same behavior, we also notice that the relative improvement of an ensemble over the neural model is bigger for rare words, as shown in Fig.~\ref{fig:rare_words}.

In order to gain better understanding, we selected sentences where the n-gram model significantly outperforms the neural model in the One Billion Word dataset, see some examples in Table~\ref{tab:examples_ngram_better}. In the vast majority of the cases, these contexts contain long proper nouns, e.g., \textit{Senator George H. Winner}.
As a quantitative evidence, we found that 23\% of words, such that $\mathrm{P}^{NG} - \mathrm{P}^{NN} > 0.5$, are  capitalized versus 13\% in general distribution.
In other cases,
we found phrases that exactly match training examples. In both cases, n-gram models are better equipped at predicting since these are essentially memorization tasks.
This also explains why n-grams yield the biggest gains on the rarest word bucket of Fig.~\ref{fig:rare_words}, despite being limited on rare words (because of data sparsity).
Overall, it seems that the task of assessing whether the n-gram model is better than the neural model is fairly easily predictable. Next, we propose a simple method to do so.

\subsection{Gating Network}

\begin{table}[]
\centering
\begin{tabular}{lr}
& \#features   \\
\hline
back-off weights ($\alpha$) & 5  \\
discounted probabilities ($p$) & 5   \\
logarithm of position ($\log(t)$) & 1  \\
\hline
$\max(\textrm{P}^{NG}(\bullet|\mathbf{c}))$, $\textrm{entropy}(\textrm{P}^{NG}(\bullet|\mathbf{c}))$ & 2 \\
$\max(\textrm{P}^{NN}(\bullet|\mathbf{c}))$, $\textrm{entropy}(\textrm{P}^{NN}(\bullet|\mathbf{c}))$ & 2 \\
\end{tabular}
\caption{ Description of features used. Top block of features are used for both SIMPLE and FULL. Lower block is used only for FULL. The notation follows that in Eq.~\ref{eq:smoothed_ngram} and Eq.~\ref{eqn:nn_lm}.}
\label{tab:feats}
\end{table}

\label{sec:gating}
A mixture of experts can be written as:
\begin{eqnarray}
\label{eq:ensemble}
    P^{ens}(w_t | \bold{c} ) = & \lambda(\bold{c} ) P^{NN}(w_t | \bold{c} ) + \nonumber \\
 & (1 - \lambda(\bold{c} )) P^{NG}(w_t | \bold{c} ),
\end{eqnarray}
where $\lambda(\bold{c})$ is a scalar between $0$ and $1$ which is the output of our gating network, and $P^{NN}$ and $P^{NG}$ are defined by Eq.~\ref{eq:smoothed_ngram} and Eq.~\ref{eqn:nn_lm} respectively.

In this work, we propose to use as gating network a small model that takes as input a handful of hand crafted features.
We choose our features to convey our intuition that switching to an n-gram model should depend on both the frequency of the word as well as the entropy of the prediction.
We therefore use both the back-off Kneser-Ney coefficients and discounted probabilities, as well as entropy of the distribution over the next word and its mode.
In order to account for positional information in the sentence (as we also found that n-grams are worse at later positions in long sentences), we also add the log of
the word position. The full list of features is given in Table~\ref{tab:feats}. We have a total of 15 features, denoted as FULL set,
which we use as input to the gating network.
We denote a subset of features that require nothing more than the coefficients from the existing n-gram model as SIMPLE.
This allows us to investigate the relative
importance of the signal from the neural language model and to further reduce the computational burden.

We train the gating network with cross-entropy loss using Eq.~\ref{eq:ensemble} as predictive distribution, and without updating the expert models.

\section{Experiments}
\label{sec:exps}
We performed language modeling at the word level using the One Billion Word benchmark~\cite{chelba2013one} and the Wall Street Journal (WSJ)~\footnote{Obtained from \url{http://www.fit.vutbr.cz/~imikolov/rnnlm/kaldi-wsj.tgz}}, see details in Table \ref{data_sets_tbl}.

\begin{table}[]
\centering
\begin{tabular}{l|rr}
                & WSJ  & 1B \\ \hline
training tokens & 36M & 768M     \\
unique words    & 20k  & 793k
\end{tabular}
\caption{Data sizes for the WSJ and 1B Word corpora.}
\label{data_sets_tbl}
\end{table}

\subsection{Expert models}\label{sec:res_experts}
We used KenLM toolkit~\citep{heafield2011kenlm} to train a 5-gram model with modified Kneser-Ney smoothing for both datasets.

For the One Billion Word dataset, we used the best neural model reported by~\citet{jozefowicz2016exploring}, composed of two LSTM layers with 8092 hidden units each, and projection layers with 1024 units. Due to the large vocabulary size, we trained using sampled softmax~\citep{jean2014using}.
We trained the model for four days on 8 GPU. For the smaller WSJ we used a much simpler model with a single LSTM layer with 500 units and trained it until convergence.
We used dropout as a regularization technique for both models~\cite{srivastava2014dropout}.

\subsection{Gating model}
The gating model was trained using only validation data. For WSJ we used 14k sentences for training and the remaining 2k for early stopping.
For 1B Word, we used two separate held-out partitions to train and validate the gating models, each with 6K sentences.
In both cases we didn't use any test set to train nor tune the gating model.

We experimented with three architectures: a linear model (LIN), a fully connected neural network with two hidden layers with 32 units each (MLP),
and an LSTM with 8 units on top of the previous MLP to test if temporal dependencies matter.

The largest gating network has only 2572 trainable parameters. All hyper-parameters were tuned on 1B Word dataset and applied to WSJ as is.
We normalize feature on the training set to have zero-mean and unit-variance.
All networks were trained until convergence with Adam optimizer~\citep{kingma2014adam} with initial learning rate of $6\times10^{-3}$, which was halved every 5k steps.

\subsection{Results}
First, we evaluated the three gating architectures  on the validation set of the One Billion Word dataset, as shown in Table~\ref{tab:ppl_valid}.
While extra features decrease the perplexity, the effect of the model architecture is much more impactful.
On the following, we only consider the LSTM architecture for the gating model.

\begin{table}[]
    \centering
\begin{tabular}{l|rr}
& \multicolumn{2}{c}{feature set } \\
model & SIMPLE & FULL \\
\hline
LIN & 30.82 & 30.75 \\
MLP & 30.35 & 30.30 \\
LSTM & 30.05 & 29.85 \\
\end{tabular}
    \caption{Validation perplexity (lower is better) for different features and architectures of the gating model. Each experiment is an average over 10 runs.}
    \label{tab:ppl_valid}
\end{table}

Next, we evaluated the model on the test sets, see Table~\ref{tab:ppl_test}.
We compare the mixture of experts with static ensembling, which is the current state of the art.

\begin{table}[]
\centering
\begin{tabular}{l|rr}
& WSJ & 1B \\
\hline
n-gram LM & 113.23 & 66.96 \\
neural LM & 71.39 & 33.01 \\
ensemble & 67.44 & 29.80 \\
HIDDEN & $67.01\pm 0.05$ & $29.72 \pm 0.03$ \\
\hline
+end-to-end & $66.88 \pm 0.07$ & - \\
\hline
\hline
SIMPLE & $67.11\pm 0.11$   & $ \mathbf{28.88}\pm 0.04$ \\
FULL & $\mathbf{66.75}\pm 0.03$ & $\mathbf{28.70} \pm 0.04$ \\
\end{tabular}
\caption{Test perplexity on WSJ and 1B (lower is better). %ADDAFTER (lower is better).}
For \textit{MoE} models we report mean and standard error over 10 runs. Results in bold show significant improvement over the best baseline (two tailed t-test, $p < 0.001$).}
\label{tab:ppl_test}
\end{table}

To compare our approach with the method proposed in~\citet{neubig2016generalizing}, we trained a gating model using the last hidden layer of the neural language model as features (HIDDEN; 500 features for WSJ and 1024 for 1B Word).
We tried two variants of the model: with and without gradient flow through neural LM.
The fine tuning requires computing of the full softmax as well as its gradient.
Attempt to do this with 1B Word neural model resulted in running out of GPU memory.
As described in~\ref{sec:res_experts} this model is trained with a sampled softmax that requires much less memory than full softmax.
Due to this reason we report results on HIDDEN features with end-to-end training only for WSJ.

On both datasets the mixture model with FULL features show a significant improvement over these baselines.
Note that on 1B Word dataset SIMPLE features alone outperform HIDDEN features by a margin.
We speculate that this is caused by the better ratio of the model size to the size of the dataset: SIMPLE features are derived from an n-gram model that scales better than a neural model.

\subsection{Conclusion}
We proposed a very simple yet effective method to combine pretrained neural and n-gram language models.
Instead of ensembling, we learn a per-time step predictor of the optimal weight between the two models.
The gating network is small, fast to train and to run, because it takes as input handcrafted features found by analyzing where n-grams outperform neural models.

\bibliography{acl2018}
\bibliographystyle{acl_natbib_nourl}

\end{document}